\documentclass[10pt,twocolumn,letterpaper]{article}

\usepackage{iccv}
\usepackage{times}
\usepackage{epsfig}
\usepackage{graphicx}
\usepackage{amsmath}
\usepackage{amssymb}

\usepackage{authblk}
\usepackage{multirow}
\usepackage{graphicx} 
\usepackage{subfigure}
\usepackage{amsmath}
\usepackage{booktabs}
\usepackage{float}
\usepackage{bm}
\usepackage{color}

\usepackage{algorithm}
\usepackage{algorithmic}

\usepackage[accsupp]{axessibility}

\usepackage[pagebackref=true,breaklinks=true,letterpaper=true,colorlinks,bookmarks=false]{hyperref}

\usepackage[breaklinks=true,bookmarks=false]{hyperref}

\iccvfinalcopy 

\def\iccvPaperID{****} 

\ificcvfinal\fi

\begin{document}

\title{Context Decoupling Augmentation for Weakly Supervised Semantic Segmentation}

\author[1,2]{Yukun Su}
\author[1,2]{Ruizhou Sun}
\author[3$^\dag$]{Guosheng Lin}
\author[1,2$^\dag$]{Qingyao Wu}
\affil[1]{School of Software and Engineering, South China University of Technology}
\affil[2]{Key Laboratory of Big Data and Intelligent Robot, Ministry of Education}
\affil[3]{School of Computer Science and Engineering, Nanyang Technological University}
\affil[ ]{suyukun666@gmail.com, ruizhousun@foxmail.com, gslin@ntu.edu.sg, qyw@scut.edu.cn}

\maketitle
\ificcvfinal\thispagestyle{empty}\fi

\renewcommand{\thefootnote}{\fnsymbol{footnote}}
\footnotetext[2]{Corresponding authors.}

\begin{abstract}
   Data augmentation is vital for deep learning neural networks. By providing massive training samples, it helps to improve the generalization ability of the model. Weakly supervised semantic segmentation (WSSS) is a challenging problem that has been deeply studied in recent years, conventional data augmentation approaches for WSSS usually employ geometrical transformations, random cropping and color jittering. However, merely increasing the same contextual semantic data does not bring much gain to the networks to distinguish the objects, $\eg$, the correct image-level classification of “aeroplane” may be not only due to the recognition of the object itself, but also its co-occurrence context like “sky", which will cause the model to focus less on the object features.
   To this end, we present a Context Decoupling Augmentation (CDA) method, to change the inherent context in which the objects appear and thus drive the network to remove the dependence between object instances and contextual information. 
   To validate the effectiveness of the proposed method, extensive experiments on PASCAL VOC 2012 and COCO datasets with several alternative network architectures demonstrate that CDA can boost various popular WSSS methods to the new state-of-the-art by a large margin. Code is available at \url{https://github.com/suyukun666/CDA}
\end{abstract}

\section{Introduction}

Semantic segmentation is a foundation in the computer vision field, which aims to predict the pixel-wise classification of the images and it enjoys a wide range of applications. Recently, benefiting from the deep neural networks,  modern semantic segmentation models~\cite{chen2017deeplab,chen2018encoder,lin2016efficient,long2015fully} have achieved remarkable progress with massive human-annotated labeled data. 
However, collecting pixel-level labels is very time-consuming and labor-intensive, which shifts much research attention to weakly supervised semantic segmentation (WSSS).
There exist various types of weak supervision for semantic segmentation like using bounding boxes~\cite{dai2015boxsup,khoreva2017simple}, scribbles~\cite{lin2016scribblesup,vernaza2017learning}, points~\cite{bearman2016s}, and image-level labels~\cite{hong2017weakly,ahn2018learning,ahn2019weakly,wang2020self,zhang2020splitting}. Among them, image-level class labels have been widely used since they demand the least annotation efforts and are already provided in existing large-scale image datasets.

\begin{figure}[]
	\begin{center}
		\centering
		\includegraphics[width=3.3in]{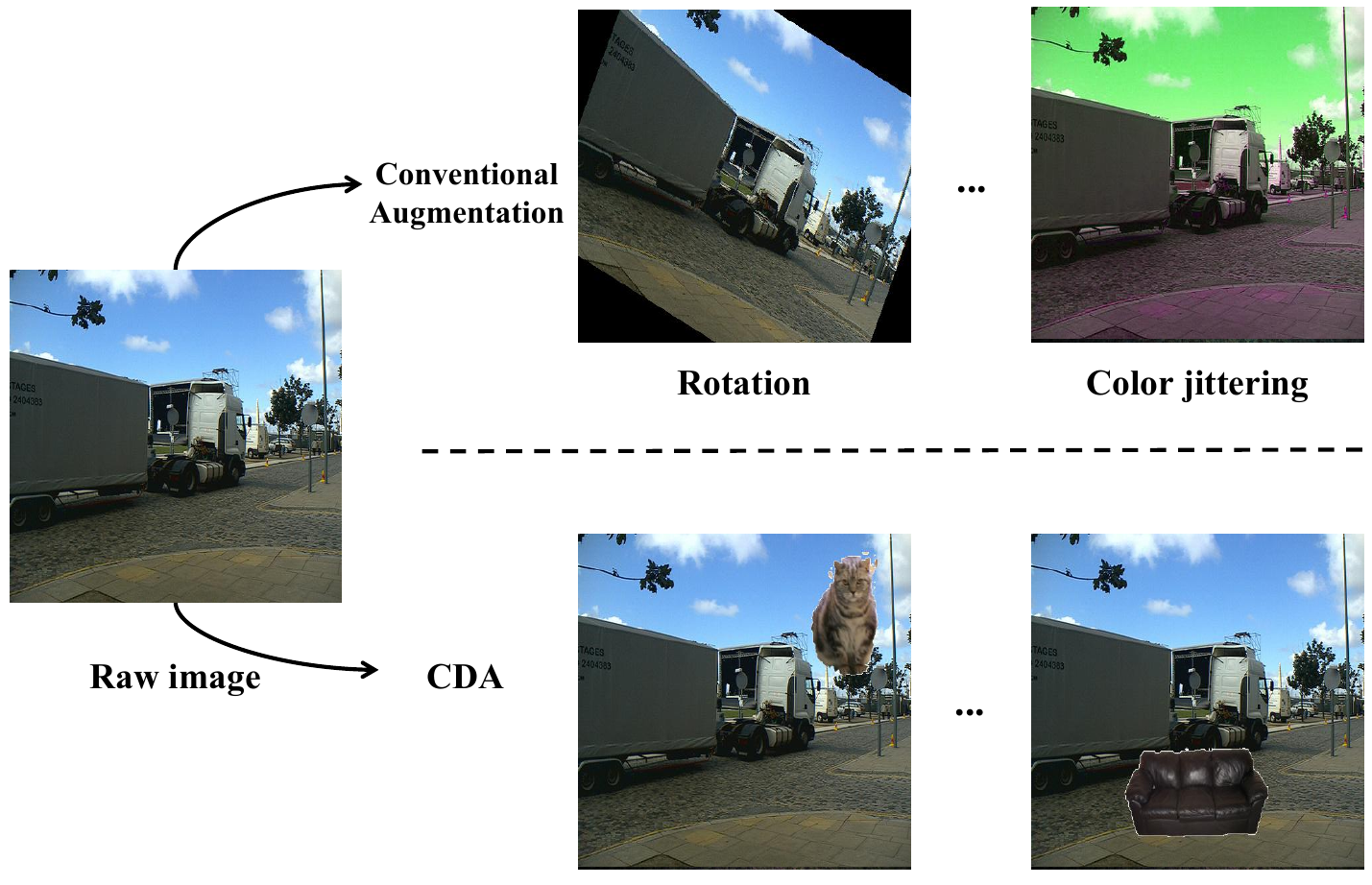}
	\end{center}
	\caption{Illustration of the difference between conventional augmentation approaches and our method. Classical data augmentation consists of generating images obtained by basic geometrical transformations or color changes of original training images. Context Decoupling Augmentation (CDA) aims to randomly paste the given object instances into the scenes, so as to decouple the inherent context position of the original objects in the image.}
	\label{fig1}
\end{figure}


In this paper, we focus on augmentation for WSSS with image-level labels, which is crucial for deep learning networks. As shown in Figure~\ref{fig1} upper part, given a training image, traditional data augmentation methods utilize some geometrical transformations, such as rotation, scaling, flipping, and even some color conversions to increase the diversity of images to avoid overfitting. However, for weakly supervised semantic segmentation, adjusting the image as a whole and maintain the same contextual semantic relation will not significantly help the networks to mine the object areas. 
For example, “sofa"  always appears in the room in the datasets, therefore, the trained network may not only recognize the objects depending on the instance features but also their co-occurrence context information~\cite{li2018tell}. 	Specifically, when object instances often appear at the same time with some accompanying backgrounds, it will cause the networks to yield confounding bias. Namely, the networks can perform classification task well is not due to successfully distinguishing the characteristics of objects, but to being aware of the appearance of certain contextual semantic information, which is harmful to mine the object regions.

Based on this observation, we propose a Context Decoupling Augmentation (CDA) method, designing for disassembling the inherent contextual information of the original image. As shown in Figure~\ref{fig1} bottom half, the “cat" shows in the “sky", and the “sofa" falls on the “road". Although some of these scene collocations rarely appear in life, the models can pay more attention to the objects corresponding to the classification labels.
Unlike the fully-supervised data augmentation approaches~\cite{dvornik2018modeling}, we cannot access the object instance labels to extract the objects under the weakly supervised setting.  
Therefore, we first adopt off-the-shelf WSSS approaches to obtain the object instances that have been well-segmented. Secondly, we randomly paste the selected foreground instances into the input images to get the new enhanced images and put them into the model for training together with  the original ones without augmentation.
In this way, we can break the dependency between objects and contextual background, and the models will focus on the internal information of the foreground instances rather than the context information to predict the categories they belong to. Besides, we use an online training technique to conduct data augmentation, which means that the combination of the raw input images and the object instances to be pasted are different each time.
This greatly increases the diversity of combinations of various scenes and object instances, and thus enhance the decoupling capability of the networks.

In the proposed context decoupling augmentation framework, we utilize different WSSS networks as our baselines. To verify the effectiveness of our proposed method, extensive experiments show that CDA can improve pseudo-masks more than 2.8\% mIoU on average. We achieve new state-of-the-art performance by 66.1\% mIoU on the $\mathit{val}$ set and 66.8\% mIoU on the $\mathit{test}$ set of PASCAL VOC 2012~\cite{everingham2015pascal}, and 33.7\% mIoU on the $\mathit{val}$ set of COCO~\cite{coco}.
The main contributions of our paper can be summarized as follows:
\begin{itemize}
	\item 
	We present a generally applicable data augmentation approach for weakly supervised semantic segmentation, which, to the best of our knowledge, has not been well explored.
	\item
	The proposed context decoupling augmentation (CDA) method does not require additional data and it can remove the correlation between foreground object instances and background context information, which can drive the network focus on object regions rather than the background.
	\item 
	Experiments on PASCAL VOC 2012 and COCO show the effectiveness of our proposed method and CDA can boost the performance of  different WSSS methods to the new state-of-the-art by a large margin.
\end{itemize}

\begin{figure*}
	\begin{center}
		\centering
		\includegraphics[width=6.0in]{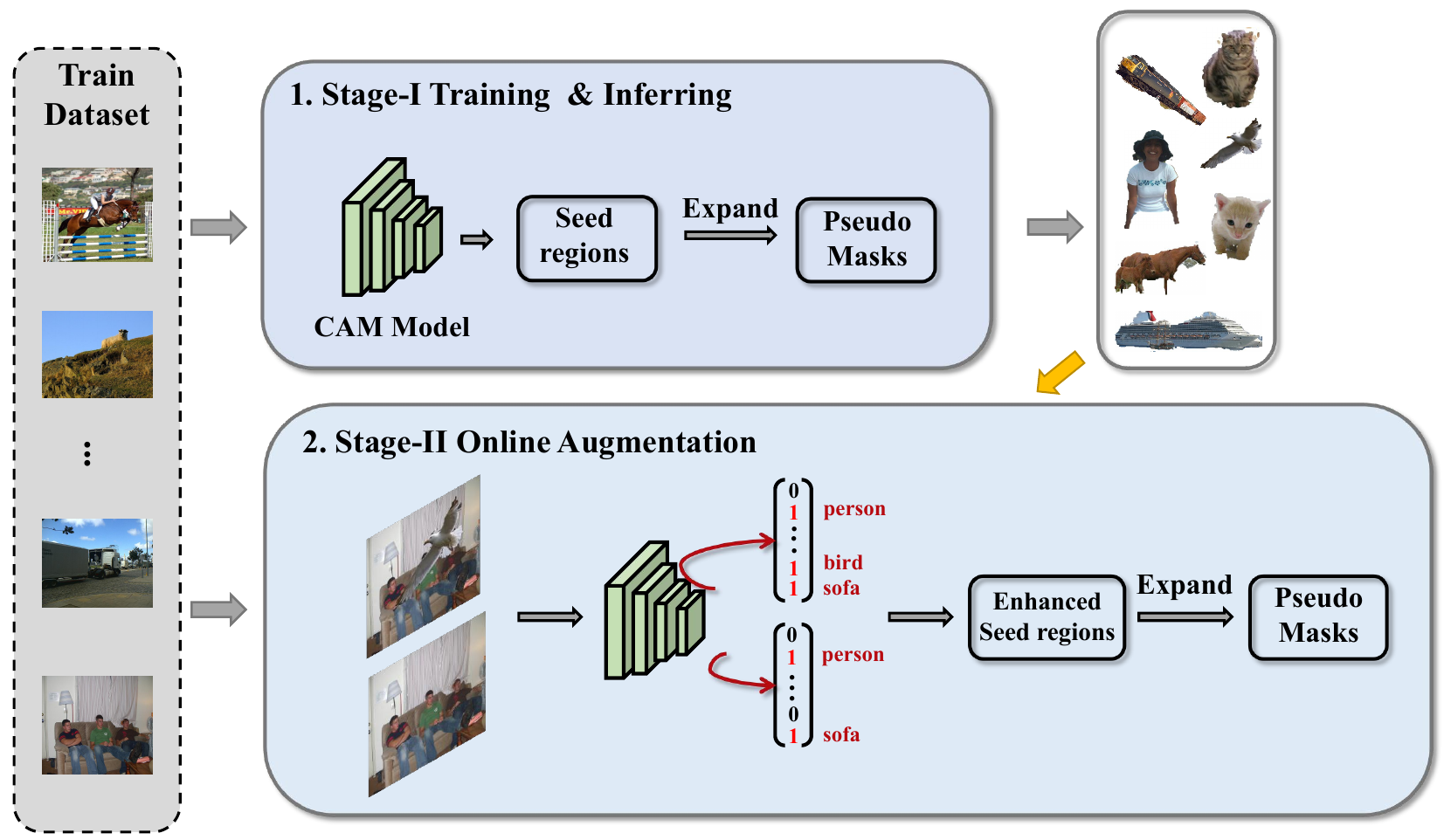}
	\end{center}
	\caption{Overview of the proposed augmentation scheme. Stage-I: use the off-the-shelf weakly supervised semantic segmentation methods to obtain some simple object instances with good segmentation. Stage-II: paste the object instances randomly into the raw images to form the new input images, and perform online data augmentation training in a pairwise way with the original input images.}
	\label{fig2}
\end{figure*}

\section{Related Work}


\subsection{WSSS}

Image labels as the weak supervision for segmentation have been widely studied in the past few years. Many approaches~\cite{wei2017object,ahn2018learning,ahn2019weakly} use CAM~\cite{zhou2016learning} to mine the object seed regions by predicting image labels. To solve the problem that only the discriminative regions can be highlighted, researchers designed to expand the object seed regions in various ways. For example, in~\cite{yu2015multi}, the target regions are expanded by fusing different discriminative regions generated by convolutional layers with different expansion rates. ~\cite{wei2017object} drives the network to learn the rest parts of the objects by iteratively erasing the target areas. In addition, some previous works~\cite{hong2017weakly,hou2018self} use additional data, such as videos and saliency maps, to explore the objects areas.

Although object expansion technologies emerge endlessly, they all use CAM~\cite{zhou2016learning} as the cornerstone. The effect of subsequent diffusion depends on the first step of the CAM learning features.
As only image-level labels are provided, when objects are closely coupled with contextual backgrounds, such as “boat" and “water", “aeroplane" and “sky", “train" and “track", CAM will mistakenly recognize the background together with foreground objects. As mentioned in~\cite{li2018tell}, the training networks have no incentive to focus attention only on the foreground class as there may be bias towards other contextual factors as a distractor with high correlation. Thus, this is an issue that's worth thinking about and that needs to be solved.

\subsection{Data Augmentation}

Data augmentation is a major trick to train deep neural networks, which aims to increase the diversity of the data by increasing the training samples and avoid overfitting to a certain extent. 
Conventional data augmentation approaches perform a series of operations on the basic data, such as rotation, flipping, adding Gaussian noise, etc.
Some works have explored synthesizing training data~\cite{frid2018synthetic,peng2015learning} for further generalizability. Generating new training samples by Stylizing ImageNet~\cite{geirhos2018imagenet} can lead to better classification performances. Recently, GAN~\cite{zhu2017unpaired} has been employed to transfer the style of the images and to make the content of the images from one domain to another, which can enrich the semantic information of the images to train the deep neural networks. 
Furthermore, ~\cite{mixup} introduced a method to mix two random samples and divide the classification results proportionally to enhance images.~\cite{cutout} conducted augmentation by randomly cutting out some areas in the sample and filled it with 0 pixel value, and keep the result of classification unchanged.

For object detection and segmentation, a popular data augmentation way is “copy-and-paste”~\cite{dvornik2018modeling,dwibedi2017cut}. These works pasted real segmented objects into natural images, which is beneficial to increase the object complexity of the internal images and can help to solve the problem of small target detection. However, obtaining these segmented objects requires pixel-wise instance labels.~\cite{remez2018learning} used box-supervision and the off-the-shelf faster-RCNN~\cite{ren2015faster} method to segment and generate masks via cut-and-paste.~\cite{arandjelovic2019object} adopted the unsupervised cut-and-paste learning method to generate new combined images, but this kind of method is only applicable to the image of single object.
It is the first time that we employ copy-and-paste in the WSSS field and it does not require the help of pixel-wise labels and other auxiliary approaches. Thus, for WSSS, such a data augmentation scheme is significant and has not been well explored.

\section{Framework}

Our approach mainly consists of two stages : (1) we first collect the easy examples of well-segmented objects by using off-the-shelf \ WSSS methods; (2) then we train the network in a pairwise manner with online augmentation. In this section, we will describe these two stages in details.

\subsection{Object Instances Collecting}

We aim to apply data augmentation on one of the WSSS models ($\ie$, IRNet~\cite{ahn2019weakly}). To some extent, the WSSS method can successfully predict good masks for some easy objects with class labels. Therefore, as shown in Figure~\ref{fig2}, in the first stage, we train the original network and we are able to select qualified object instances through the scene complexity of the image, the scope of the object and the semantic relevance by setting some criteria. 

Specifically, for the inferring phase after training the network, we follow two main criteria for collecting object instances: (i) the current image should only have a single class. The intuition behind this is that in the case of only a single class, the image information should be simple and without a complex semantic environment, the segmentation results of the model should be more accurate; (ii) the segmentation result of the current image should meet the condition, $\epsilon_1 < \frac{m}{n} < \epsilon_2$, where $\epsilon_1$ and $\epsilon_2$ are two threshold factors, respectively. $m$ is the number of pixels belonging to the foreground object, $n$ is the number of pixels of the entire image. The reason lies that if the scale value of $\frac{m}{n}$ is too large, it should be that the background is incorrectly identified as the foreground. In contrast, if the scale value is too small, it should be that the model has not been able to recognize enough foreground object pixel information.
Different from existing synthesis approaches~\cite{dvornik2018modeling,dwibedi2017cut} , our method is based on self-provided masks to obtain qualified object instances images.

\begin{figure}[]
	\begin{center}
		\centering
		\includegraphics[width=3.3in]{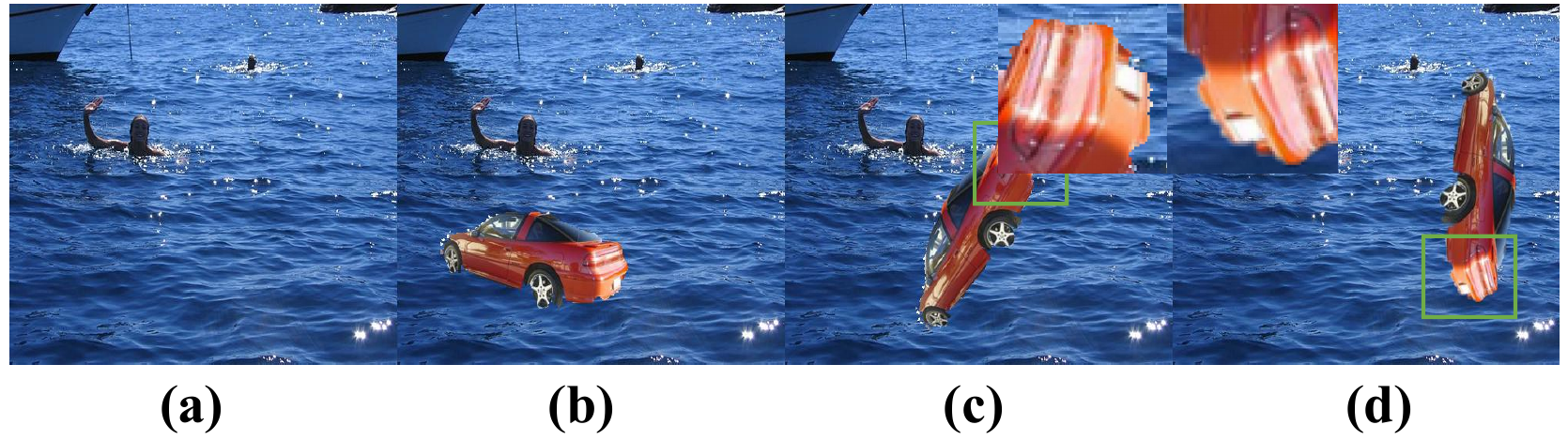}
	\end{center}
	\caption{Different kinds of pasting methods used in experiments. (a) Raw input, (b) Random rescale pasting, (c) Random rescale + rotation pasting, (d) Random rescale + rotation +  Gaussian smoothing pasting. }
	\label{fig3}
\end{figure}

\subsection{Online Augmentation Training}

\textbf{Blending.} 
Before we take a step to train the network in the second stage, we first introduce how to blend the object instances into the natural images.
As shown in Figure~\ref{fig3}, we show different types of pasting skills in our experiments. It’s worth mentioning that we only paste objects that have not appeared in the original images. The significance of this is that we can increase the diversity of objects of the images, while also reducing the dependence of the same objects in the inherent scene. 
By randomly rescaling the objects, we can paste them into the images appropriately to prevent them from being too large or too small. The addition of random rotation can change the inherent orientation properties of the objects. Adding Gaussian smoothing can help the added objects boundary blend more naturally.

\begin{figure}[]
	\begin{center}
		\centering
		\includegraphics[width=3.3in]{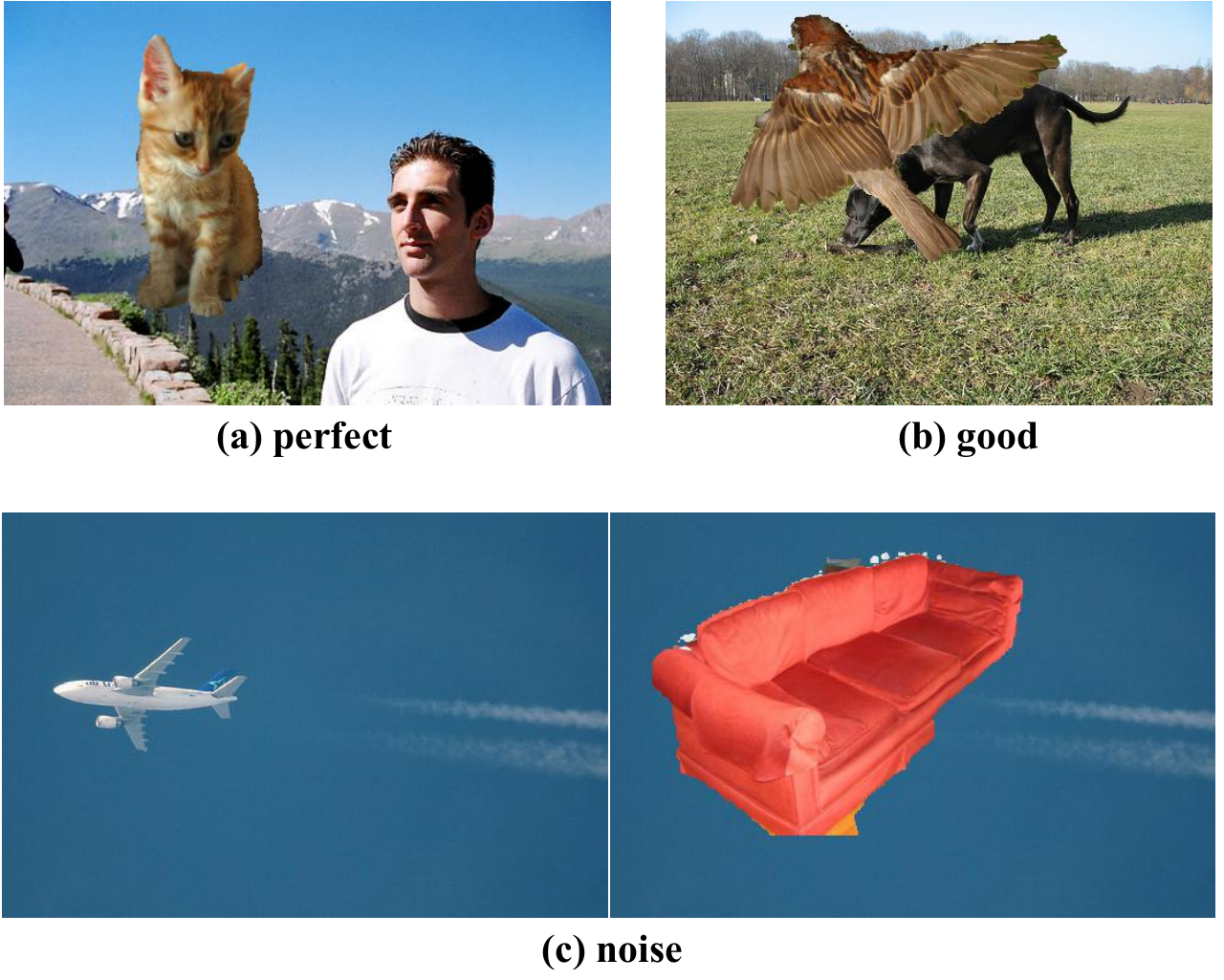}
	\end{center}
	\caption{Examples of the input augmented images with varying degrees of occlusion.}
	\label{fig4}
\end{figure}

In some cases, the blending may not be ideal, we elaborate on several possibilities for random pasting. As shown in Figure~\ref{fig4}, we have listed several augmented images of random pasting and we call them “perfect", “good" and “noise" examples.
As for the “good" example, the new object “bird"  covers part of the “dog” in the original image, however, we argue that this could help to erase the discriminative regions and force the network to discover more object regions like the function in ~\cite{wei2017object}. The noise example shows that the “sofa" completely covers the “aeroplane” in the original image, which will cause confusion to network classification. 
However, we consider that such hard examples do not account for the majority. Most objects occupy in the middle or prominent location of the natural images. The random blending method we employ tends to paste the new objects into the off-center position of the images. Thus, this case does not affect learning. Hence, our framework is robust to the quality of augmentation. According to our experiments, this simple random blending method performs well in boosting the performance.

\textbf{Online Training.} 
The augmentation scheme is conducted online to enhance the network trained in stage-I to improve the ability to distinguish object features.
Formally, in each batch, we sample $N/2$ images from the training dataset and the same number object instances images  from the subset which is provided from stage-I. Then we randomly paste the segmented objects into the input images, which creates a $N/2$ batch new images. Thus, a batch of size $N$ is generated online for each augmentation iteration.
The construction process of the online augmentation learning is summarized in Algorithm~\ref{alg:Framwork}.

\renewcommand{\algorithmicrequire}{ \textbf{Input:}} 
\renewcommand{\algorithmicensure}{ \textbf{Output:}} 

\begin{algorithm}[htb] 
	\caption{Stage-II: Online Augmentation.} 
	\label{alg:Framwork} 
	\begin{algorithmic}[1] 
		\REQUIRE ~~\\ 
		The training dataset images $\mathcal{I}$ and the corresponding labels $\mathcal{L}$;\\
		The object instances $\mathcal{O}$ and the corresponding labels $\mathcal{T}$.
		\label{ code:fram:extract }
		\WHILE{not done} 
		\STATE ($\mathcal{I}_i$, $\mathcal{L}_i$) $\gets$ Draw one sample from training dataset;\
		\STATE ($\mathcal{O}_j$, $\mathcal{T}_j$) $\gets$ Draw one sample from object instances subset;\
		\WHILE{$\mathcal{T}_j$ in $\mathcal{L}_i$} 
		\STATE ($\mathcal{O}_j$, $\mathcal{T}_j$) $\gets$ Resample;\
		\ENDWHILE
		\STATE $\mathcal{I}^{'}_i$ $\gets$ Blend $\mathcal{O}_j$ into $\mathcal{I}_i$;\
		\STATE $\mathcal{L}^{'}_i$ $\gets$ Append $\mathcal{T}_j$ in $\mathcal{L}_i$;\
		\STATE Train CAM $\gets$ Loss($\mathbb{C}$($\mathcal{I}_i$), $\mathcal{L}_i$) + Loss($\mathbb{C}$($\mathcal{I}^{'}_i$), $\mathcal{L}^{'}_i$);\
		\ENDWHILE
		\label{code:fram:trainbase}
		\STATE Expansion. 
	\end{algorithmic}
\end{algorithm}

Note that we train the online augmentation method in a pairwise manner as shown in Figure~\ref{fig2} stage-II left. We consider this can further help the networks to recognize the objects for the reason that some images have new blended objects, while some do not, which can help the classifier find more discriminative features. 
The motivation behind this is similar to “finding the differences" with the human visual system. When the two images have a different object but with a duplicated background, which can often leave a deep impression. For the same reason, this can make the network classifier learn better features of this kind of object.

\subsection{Discussion}

The proposed CDA framework contributes a new data augmentation learning strategy. Unlike the previous “copy-and-paste” works, we do not use additional pixel-wise labels.
Specifically, by using the self-provided initial segmentation masks of the models, we can obtain the object instances for the next phase augmentation training. Furthermore, since our goal is to decouple the high correlation between objects and their contextual background, we don’t need to consider much about visual context~\cite{dvornik2018modeling,chu2018deep}, which can greatly improve the efficiency of pasting objects into the images.
Besides, we adopt online augmentation training skills. Compared with static offline data augmentation, which merely enlarges the scale of the training dataset in linear-level. Namely, once a new dataset is formed, the number of images will remain unchanged. However, our method is able to obtain exponential-level augmentation, because the combination of object instances and natural images can be ever-changing in each round of training.

\section{Experiments}

To demonstrate the contributions of the proposed method, we conduct several ablation studies to show the effectiveness of CDA and compare different baselines models to the state-of-the-arts. We will give the details of the datasets, evaluation metric, and baseline models in the following.

\subsection{Dataset}

All the networks in our framework are trained and evaluated on the PASCAL VOC 2012~\cite{everingham2015pascal} and COCO ~\cite{coco} segmentation benchmark for a fair comparison to previous approaches.
As for PASCAL VOC, the official dataset separation has 1464 images for training, 1449 for validation and 1456 for testing.
Following the common practice, we take additional annotations to build an augmented training set with 10582 images presented in~\cite{hariharan2011semantic}. COCO is a more challenging benchmark with 81 semantic classes (one background class), 80k, and 40k images for training and validation.
We use the standard mean Intersection-over-Union (\textbf{mIoU}) as the evaluation metric for all experiments.

\subsection{Implementation Details}

To validate the applicability of CDA, we deploy it on three popular WSSS models including IRNet~\cite{ahn2019weakly}, AffinityNet~\cite{ahn2018learning} and SEAM~\cite{wang2020self}. 
The general training architecture components include a multi-label image classification step, a
pseudo-mask generation step, and the final segmentation model (DeepLab-v2~\cite{chen2017deeplab}). We strictly follow the same settings as reported in the official codes. Specially, for SEAM~\cite{wang2020self} and AffinityNet~\cite{ahn2018learning} baselines, ResNet38~\cite{he2016deep} that pre-trained on ImageNet~\cite{deng2009imagenet} is adopted as backbone with batch size as 8 and 16, respectively. When training the networks,  multi-scale and data augmentation techniques like horizontal flip, random cropping, and color jittering are deployed in both architectures. Following the poly policy $lr_{init}$ = $lr_{init}(1-itr/max\_itr)^\rho$ with $\rho$ = 0.9 for decay, the models are trained with a fix input size as 448 $\times$ 448 using Adam optimizer~\cite{kingma2014adam}. Besides, online hard example mining~\cite{shrivastava2016training} is employed on the training loss in SEAM.
As for IRNet~\cite{ahn2019weakly}, ResNet50~\cite{he2016deep} is used as the backbone network (pretrained on ImageNet). The batch size is set to 16 for the image classification model and 32 for the inter-pixel relation model. The input image is cropped into a fix size of 512 $\times$ 512 using zero padding if needed. The model is trained with the same polynomial decay strategy as in AffinityNet~\cite{ahn2018learning} using stochastic gradient descent (SGD) for optimization with 8, 000 iterations. 
The fully-connected CRF~\cite{krahenbuhl2011efficient} is used in three baselines to refine CAM, pseudo-mask, and segmentation mask with the default parameters in the public code.
We set the threshold $\epsilon_1$ = 0.1 and $\epsilon_2$ = 0.7 by experience.

\subsection{Ablation Studies}

To verify the effectiveness of our CDA, we evaluate CAM seed regions, pseudo-masks, and segmentation masks, respectively. In our experiments, the standard mean Intersection over Union (mIoU) is used on the training set for evaluating CAM seed area masks and pseudo-masks, and on the PASCAL VOC 2012 $\mathit{val}$ and $\mathit{test}$ sets for evaluating segmentation masks. For the sake of simplicity, since the three WSSS models are all based on CAM~\cite{zhou2016learning}, we use one of the representative models (IRNet~\cite{ahn2019weakly}) as a baseline to conduct several ablation studies on CAM in mIoU to illustrate the role of each component of our approach.

\begin{table}[]
	\begin{center}
		\scalebox{0.9}{
			\begin{tabular}{c|c|c}
				\toprule  
				\toprule  
				Method  \ & \ operation \ & \ mIoU (\%)\\
				\midrule  
				\multirow{2}*{Conventional Augmentation }& Rotation & 48.5  \\
				& Translation & 48.4 \\
				\midrule  
				\multirow{3}*{Mixup~\cite{mixup}}& $\alpha$ = 0.3& 48.7  \\
				& $\alpha$ = 0.5 & 48.5 \\
				& $\alpha$ = 0.8 & 49.0 \\
				\midrule  
				CutOut~\cite{cutout} & Random& 48.9 \\
				\midrule  
				CutMix~\cite{cutmix} &  Random& 49.2 \\
				\midrule  
				Random pasting (ours) & Rescale & \textbf{49.8} \\
				\bottomrule 
		\end{tabular}}
	\end{center}\caption{Experiments of different augmentation methods. Here $\alpha$ is the intensity of the interpolation between the eigenvector and the target vector.}\label{table1}
\end{table}

\begin{table}[]
	\begin{center}
		\scalebox{0.9}{
			\begin{tabular}{ccccc}
				\toprule  
				\toprule  
				Baseline  & Rescale &Rotation& Gaussian & mIoU (\%)\\
				\midrule  
				$\checkmark$&  & & & 48.3 \\
				$\checkmark$&  $\checkmark$& & & 49.8 \\
				$\checkmark$&  $\checkmark$&$\checkmark$ & & \textbf{50.8} \\
				$\checkmark$&  $\checkmark$& & $\checkmark$& 49.6 \\
				$\checkmark$&  $\checkmark$&$\checkmark$ & $\checkmark$& 50.4 \\
				\bottomrule 
		\end{tabular}}
	\end{center}\caption{The ablation study of the effect on different pasting methods. Baseline indicates the original CAM method without pasting new objects for augmentation.}\label{table2}
\end{table}

\begin{table}[]
	\begin{center}
		\scalebox{0.9}{
			\begin{tabular}{c|c}
				\toprule  
				\toprule  
				Training manner  \ & \ \ mIoU (\%) \\
				\midrule  
				Pairwise& \textbf{50.8} \\
				None-pairwise & 50.1 \\
				\bottomrule 
		\end{tabular}}
	\end{center}\caption{Experiments of augmentation training manner.}\label{table3}
\end{table}

\textbf{Random pasting vs. Other sophisticated augmentation methods:} As for the traditional augmentation methods, we adopt the random rotation and translation to expand the dataset to three times the original size, however, they can not bring significant boost for the performance. We also compare Mixup~\cite{mixup}, CutOut~\cite{cutout} and CutMix~\cite{cutmix} methods to generate new augmented images.
As shown in Table~\ref{table1}, random rescale pasting outperforms the other three methods achieving \textbf{49.8}\% mIoU. 
These results demonstrate that random pasting is suitable for our CDA framework. We consider that proper occlusion helps the network to better mine the features of other areas of the objects, and the situation of complete occlusion is relatively rare which will not affect our learning process. 

\begin{figure}[]
	\begin{center}
		\centering
		\includegraphics[width=3.3in]{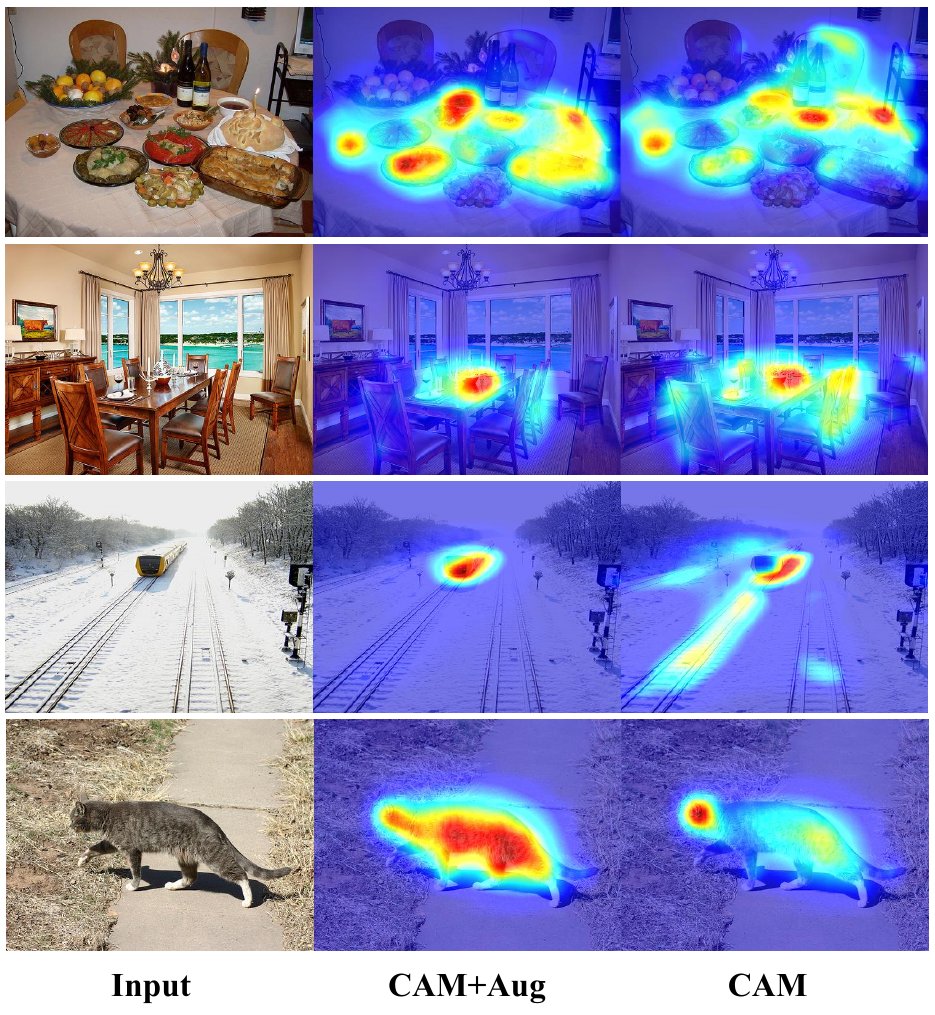}
	\end{center}
	\caption{Qualitative visualization of CAMs. Our CDA framework not only suppresses over-activation ($1^{st}, 2^{nd}, 3^{rd}$ row) of the high correlation contextual backgrounds of the objects and expands CAMs to cover the whole object regions ($4^{th}$ row).}
	\label{fig5}
\end{figure}

\textbf{Comparison with baseline:} We further explore the impact of different pasting methods on data augmentation. Table~\ref{table2} shows that using random rescale pasting has a 1.5\% improvement compared to baseline. After combining rescale and rotation, we can get the best performance to \textbf{50.8}\% mIoU on PASCAL VOC training set. The results show that applying Gaussian smoothing can not help to improve the performance. Therefore, in subsequent experiments, unless otherwise specified, we will use the random rescale combining with the rotation method.

\begin{table*}[]
	\begin{center}
		\scalebox{0.9}{
			\begin{tabular}{c|c|c|c|c|c}
				\toprule  
				\toprule  
				Network  & \  \ Backbone \ & \ \ \ \ CAM \ \ \ \ & \ Pseudo-Masks \ & \ Seg. Masks (val-set)  & \ Seg. Masks (test-set)\\
				\midrule  
				AffinityNet~\cite{ahn2018learning} &  ResNet-38 & 48.0 & 59.7  & 61.7 & 63.7\\
				+ CDA &  ResNet-38 & $48.9_{\color{red}+0.9}$ & $63.3_{\color{red}+3.6}$  & $64.2_{\color{red}+2.5}$  & $65.8_{\color{red}+2.1}$\\
				\midrule  
				IRNet$^{*}$~\cite{ahn2019weakly} &  ResNet-50 & 48.3 & 65.9  & 63.5 & 64.8\\
				+ CDA &  ResNet-50 & $50.8_{\color{red}+2.5}$ & $67.7_{\color{red}+1.8}$  & $65.8_{\color{red}+2.3}$  & $66.4_{\color{red}+1.6}$\\
				\midrule  
				SEAM~\cite{wang2020self} &  ResNet-38 & 55.4 & 63.4  & 64.5 & 65.7\\
				+ CDA &  ResNet-38 & $58.4_{\color{red}+3.0}$ & $66.4_{\color{red}+3.0}$  & $66.1_{\color{red}+1.6}$  & $66.8_{\color{red}+1.1}$\\
				\bottomrule 
		\end{tabular}}
	\end{center}\caption{Different baselines with our CDA framework performance in mIoU on PASCAL VOC. $^{*}$denotes our reimplemented results since the original code does not provided pre-trained weights.}\label{table5}
\end{table*}

Figure~\ref{fig5} shows the qualitative comparison between our CAM+Aug by CDA method and the original CAM.
As shown in the first and second rows in the figure and the labels of objects are “table". The original CAM will activate background semantic information that is strongly related to the “table”, such as “chair”. However, by employing  the decoupling augmentation training strategy, our method can focus on the target areas. For the image with the label of “train”, CAM even pays attention not to the object itself, but the “track”, which will be detrimental to the subsequent segmentation task. Moreover, CDA can also help the network expand and discover more comprehensive object features but not only the most discriminative regions like the “cat” shown in the last row.

\textbf{The effect on pairwise training:} Compared to merely using the augmented images to train the networks, we use the none-augmented images with the augmented images as pair images to jointly train the models as shown in Figure~\ref{fig2} stage-II.
The results shown in Table~\ref{table3} show that applying pairwise training strategy outperforms the one in single augmented images, which illustrates that this helps the network classifier to learn more discriminative features.

\begin{table}[]
	\begin{center}
		\scalebox{0.85}{
			\begin{tabular}{c|c|c}
				\toprule  
				\toprule  
				Number of pasted objects  &  Same category objects  & mIoU (\%)\\
				\midrule  
				1 & $\times$ & \textbf{50.8}\\
				2 & $\times$ & 48.9\\
				3 & $\times$ & 47.8\\
				\bottomrule 
				1 & \checkmark & 50.2\\
				2 & \checkmark & 48.6\\
				3 & \checkmark & 47.4\\
				\bottomrule 
		\end{tabular}}
	\end{center}\caption{Experiments of different number of pasted objects for augmentation.}\label{table4}
\end{table}


\textbf{The effect on objects numbers:} Under the default settings of our experiment, we only paste one new instance that does not exist in the original images. We further explore the effect of pasting multiple objects into the images to conduct augmentation. As shown in Table~\ref{table4} above the solid line, when the number of object to be pasted increases from one to two, the mIoU performance will decrease. As the number of pasted objects changes to three, it will even worse than the baseline. The results show that over-pasted objects may cover the objects in the original image, making the noise sample dominant. This will confuse the classifier, which will bring negative effects. 
In addition, as depicted below the solid line in Table~\ref{table4}, when we allow the pasted object to be consistent with the object category in the original image, their general performance is worse than the former. 
This shows that forcing objects of different categories to be pasted into images can decouple the strong contextual dependence of objects in the original semantic environment.

\begin{figure}[]
	\begin{center}
		\centering
		\includegraphics[width=3.3in]{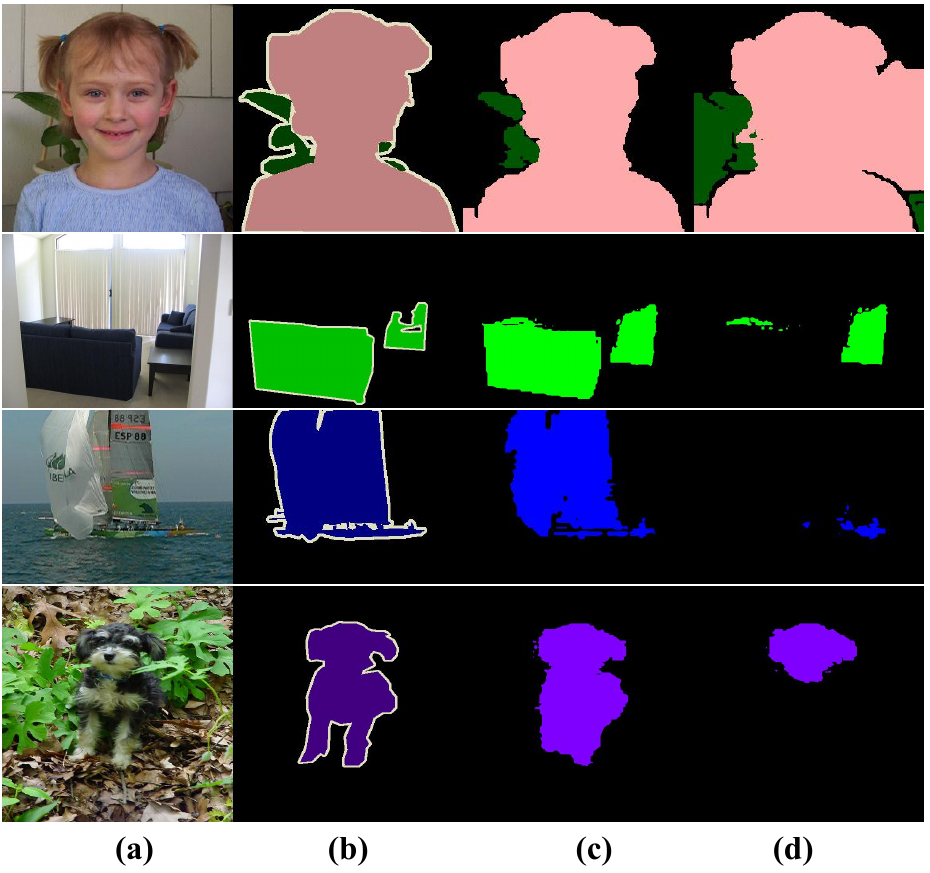}
	\end{center}
	\caption{Visualization of pseudo-masks (baseline: IRNet~\cite{ahn2019weakly}). (a) Input images. (b) Ground-Truth labels. (c) Our CAM+Aug. (d) Original CAM.}
	\label{fig6}
\end{figure}

\begin{figure*}
	\begin{center}
		\centering
		\includegraphics[width=6.0in]{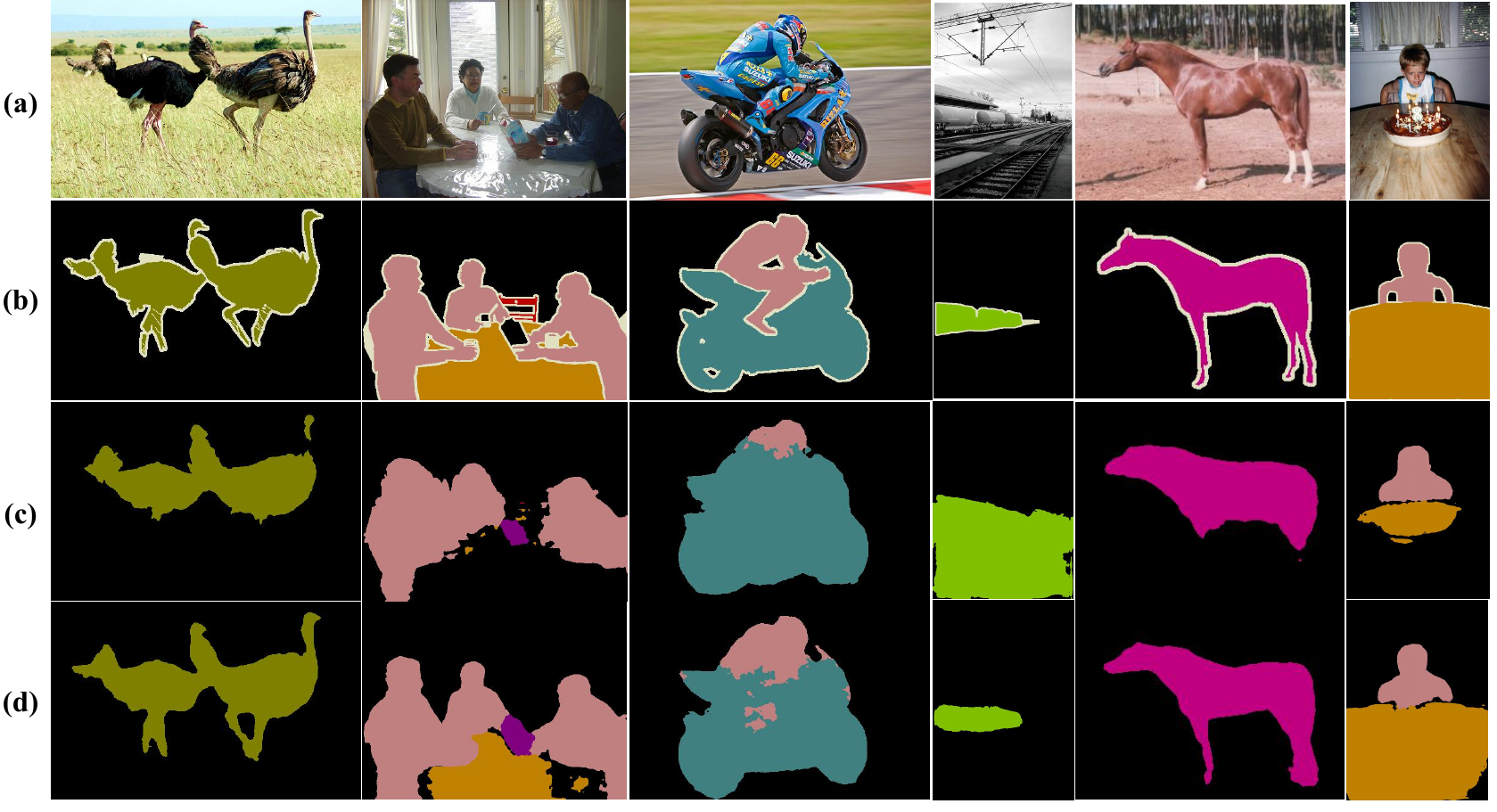}
	\end{center}
	\caption{Qualitative results on the PASCAL VOC 2012 val set. (a) Input images. (b) Ground-truth labels. (c) Results obtained by IRNet~\cite{ahn2019weakly} baseline. (d) Results of our IRNet + CDA. More results can be found in the supplementary material.}
	\label{fig7}
\end{figure*}

\begin{table}[]
	\begin{center}
		\scalebox{0.88}{
			\begin{tabular}{ccc|cc}
				\toprule  
				\toprule  
				Methods & Backbone &  Saliency & $\mathit{val}$ & $\mathit{test}$\\
				\midrule  
				CCNN~\cite{pathak2015constrained}$_{\text{ICCV'15}}$ & VGG16 & -& 35.3 & 35.6\\
				SEC~\cite{kolesnikov2016seed}$_{\text{ECCV'16}}$ & VGG16 & -&50.7& 51.1\\
				STC~\cite{wei2016stc}$_{\text{TPAMI'17}}$ & VGG16 & \checkmark&49.8 & 51.2\\
				AdvEra~\cite{wei2017object}$_{\text{CVPR'17}}$& VGG16 & \checkmark & 55.0 & 55.7\\
				DCSP~\cite{chaudhry2017discovering}$_{\text{BMVC'17}}$& ResNet101 & \checkmark & 60.8 & 61.9\\
				MDC~\cite{wei2018revisiting}$_{\text{CVPR'18}}$& VGG16 & \checkmark & 60.4 & 60.8\\
				MCOF~\cite{wang2018weakly}$_{\text{CVPR'18}}$& ResNet101 & \checkmark & 60.3 & 61.2\\
				DSRG~\cite{huang2018weakly}$_{\text{CVPR'18}}$& ResNet101 & \checkmark & 61.4 & 63.2\\
				AffinityNet~\cite{ahn2018learning}$_{\text{CVPR'18}}$& ResNet-38 & - & 61.7 & 63.7\\
				IRNet~\cite{ahn2019weakly}$_{\text{CVPR'19}}$& ResNet50 & - & 63.5 & 64.8\\
				FickleNet~\cite{lee2019ficklenet}$_{\text{CVPR'19}}$& ResNet101 & \checkmark & 64.9 & 65.3\\
				SEAM~\cite{wang2020self}$_{\text{CVPR'20}}$& ResNet38 & - & 64.5 & 65.7\\
				ICD~\cite{IDC}$_{\text{CVPR'20}}$ & ResNet101 & - & 64.1 & 64.3 \\
				\midrule  
				IRNet + CDA (ours)& ResNet50 & - & {\color{blue} 65.8} & {\color{blue} 66.4} \\
				SEAM + CDA (ours)& ResNet38 & - & {\color{red} 66.1} & {\color{red} 66.8} \\
				\bottomrule 
		\end{tabular}}
	\end{center}\caption{Performance comparisons with other state-of-the-art WSSS methods on PASCAL VOC 2012 dataset. The {\color{red} best} and {\color{blue} second best} performance under each set are marked with corresponding formats.}\label{table6}
\end{table}

\begin{table}[]
	\begin{center}
		\scalebox{0.9}{
			\begin{tabular}{c|c|c}
				\toprule  
				\toprule  
				Methods  \ & \ Backbone \ & \ $\mathit{val}$\\
				\midrule  
				BFBP~\cite{saleh2016built}$_{\text{ECCV'16}}$ & VGG16 & 20.4 \\
				SEC~\cite{kolesnikov2016seed}$_{\text{ECCV'16}}$ &  VGG16& 22.4 \\
				IRNet~\cite{ahn2019weakly}$_{\text{CVPR'19}}$& ResNet50  & 32.6  \\
				SEAM~\cite{wang2020self}$_{\text{CVPR'20}}$ & ResNet38  & 31.9  \\
				IAL~\cite{wang2020weakly}$_{\text{IJCV'20}}$ & VGG16 & 27.7\\
				\midrule  
				IRNet + CDA (ours)& ResNet50  & {\color{red} 33.7}  \\
				SEAM + CDA (ours)& ResNet38  & {\color{blue} 33.2}  \\
				\bottomrule 
		\end{tabular}}
	\end{center}\caption{Performance comparisons with other state-of-the-art WSSS methods on
	COCO val in terms of mIoU.}\label{table7}
\end{table}

\textbf{Analysis of pseudo labels and Segmentation masks:} The overall results are shown in Table~\ref{table5}. We can observe that deploying CDA on different weakly supervised semantic segmentation models can improve all their performances. Specifically, SEAM~\cite{wang2020self}  can achieve the best performance in Segmentation Masks on both validation set and testing set.
Figure~\ref{fig6} shows that we can obtain more accurate and complete masks covering the object areas. 

\subsection{Comparison with State-of-the-arts}

Finally, we compare our framework with state-of-the-art methods on the PASCAL VOC 2012 and COCO dataset including both the validation set and the testing set. For a fair comparison, we adopt the same DeepLab~\cite{chen2014semantic,chen2017deeplab} architectures as reported in the original papers.
On PASCAL VOC 2012, as is shown in Table~\ref{table6}, although different baselines already boosts performance compared to previous methods, when CDA is deployed in the models, SEAM~\cite{wang2020self} can achieve the best performance and outperform other state-of-the-arts by a large margin. IRNet~\cite{ahn2019weakly} yield the second best performance and can beat its later published works. On COCO, CDA deployed on IRNet achieves 33.7\% mIoU on the val set, which surpasses the previous best model by 1.1\% mIoU.
Figure~\ref{fig7} presents qualitative results of our CDA approach applying on IRNet baseline and compares them to itself.
We can observe that CDA can make more accurate predictions on objects, which shows better demarcations in some coherent areas. Meanwhile, CDA can help to expand and discover more comprehensive object regions. 

\vspace{10pt}
\section{Conclusion}

In this paper, we propose a Context Decoupling Augmentation (CDA) method for WSSS and to narrow the gap with fully supervision.
Specifically, through a two-stage training, the object instances provided by the network itself are copied and pasted into the input images to conduct augmentation. To further improve the ability of network for learning object features, we adopt pairwise training manner to help the classifier to distinguish more discriminative features.
Experimental results show that CDA can help boost various WSSS methods to the new state-of-the-arts.

\section*{Acknowledgement}

This work was supported by National Natural Science Foundation of China (NSFC) 61876208, Key-Area Research and Development Program of Guangdong Province 2018B010108002, Central Universities of China under Grant D2192860, and the National Research Foundation, Singapore under its AI Singapore Programme (AISG Award No: AISG-RP-2018-003), and the MOE Tier-1 research grants: RG28/18 (S), RG22/19 (S) and RG95/20.

{\small
\bibliographystyle{ieee_fullname}
\bibliography{egbib}
}

\end{document}